%% file: main_arxiv.tex
\documentclass{article}

\usepackage[usenames,dvipsnames,svgnames]{xcolor}

\usepackage{hyperref}
\hypersetup{
    colorlinks=true,
    citecolor=MidnightBlue,
    urlcolor=Bittersweet,
}

\usepackage[margin=1in]{geometry}

\include{header.tex}

\usepackage{caption}
\usepackage{subcaption}

\numberwithin{equation}{section}

\author{%
  Zhishen Huang \\
  Amazon\\
  Seattle, WA 98109 \\
  \texttt{zhishen.huang@colorado.edu} 
}

\title{Reinforcement Learning for Sampling on Temporal Medical Imaging Sequences}
\begin{document}
\maketitle
\begin{abstract}
Accelerated magnetic resonance imaging resorts to either Fourier-domain subsampling or better reconstruction algorithms to 
deal with fewer measurements while still generating medical images of high quality.
Determining the optimal sampling strategy given a fixed reconstruction protocol often has  combinatorial complexity. In this work, we apply double deep Q-learning and REINFORCE algorithms to learn the sampling strategy for dynamic image reconstruction. We consider the data in the format of time series, and the reconstruction method is a pre-trained autoencoder-typed neural network (U-Net). We present a proof of concept that reinforcement learning algorithms are effective to discover the optimal sampling pattern which underlies the pre-trained reconstructor network (i.e., the dynamics in the environment).
The code for replicating experiments can be found at \url{https://github.com/zhishenhuang/RLsamp}.
\end{abstract}
\section{Introduction}
Magnetic resonance imaging (MRI) is a non-radioactive, non-invasive medical imaging process to visualize human organs and tissues for diagnostic purposes. MRI scanners sequentially collect measurements in the frequency domain (or $k$-space), from which an image is reconstructed. A central challenge in MRI is its time-consuming sequential acquisition process as the scanner needs to densely sample the underlying $k$-space for accurate reconstruction, and during the scanning process patients are expected to remain still. The physical nature of MRI process makes it difficult to image objects in motion such as hearts, compromises image quality by motion artifacts, and incurs patients' discomfort and safety concerns. Hence, reconstructing high-quality images from limited measurements is desirable. There are two core parts in the accelerated MRI pipeline: a sampling pattern deployed to collect the (e.g., limited/undersampled) data in $k$-space and a corresponding reconstruction method (reconstructor) that also enables recovering any missing information. 
Reconstruction methods can be motivated by mathematical theory of compressed sensing~\cite{candes_cs,Becker2011PracticalCS} or data-driven methodology. Several deep learning based approaches~\cite{Sai_recon_review,transform_learning_Sai,unrolling_eldar,huang2021modelbased} have been shown to demonstrate high reconstruction quality and robust performance to various sources of artifacts and hyperparameter settings.

The experimental design of subsampling schemes is
more difficult to determine due to the combinatorial nature of this optimization problem. For a set of images $\{\x_i\}_{i=1}^N$ where $\x_i \in \R^{m\times n}$, we aim to find the optimal sampling pattern in the $k$-space to cater for the best reconstruction outcome. Let $\samp$ be the operation of applying a sampling mask on the $k$-space information, $\recon$ be the operator of reconstructing the image from $k$-space measurements, and $\fft[\x]$ be the 2D-Fourier transform. With a choice of loss function $\mathcal{L}(\cdot,\cdot)$, the optimization problem is formulated as
\begin{align}
\label{eqn::opt_prob_general}
    &\min_{\samp}\sum_{i=1}^N \mathcal{L}( \recon \circ \samp \circ \fft[\x_i] \, , \, \x_i) \nonumber\\ 
    &\textrm{s.t. }\samp \textrm{ satisfies designated sampling budget}
\end{align}

A general sampling operator $\samp$ can vary with respect to each input image $\x_i$. As a simple example, one family of sampling pattern for medical images is one-dimensional Cartesian sampling in $k$-space. With one-dimensional sampling mask, the sampling takes place with respect to one dimension of the input image and meanwhile the other dimension is fully sampled. This sampling pattern can be coded into MRI machines, and its performance is corroborated by the compressed sensing theory. If the sampling happens to the second dimension, then $\samp \circ \fft[\x] = \mathrm{FFT}[\x] \cdot \mathrm{diag}(\mathbf{m})$, where  $\mathbf{m}\in \{0,1\}^n$ is a binary vector, $\mathrm{diag}(\mathbf{m})$ is an $n\times n$ matrix with $\mathbf{m}$ on its diagonal. The sampling budget in this case can be phrased as $\sum_{i=1}^n m_i \le b$ for some $b\in \mathbb{N}^+$.

In this work, we consider the data in the form of a time series $\{\x_t\}$. We aim to build a sampling policy for a series of images that contains time evolution information. Specifically, the sampling policy is realized through a policy function $\pi(\cdot| \widehat{\x}_{t_{k-\alpha}:t_k} )$ 
given reconstruction history in the recent $\alpha$ steps. As $t$ increases, the sampling policy $\pi$ informs the next frequency location to sample. This is different from precedent study on adaptive sampling where the target under investigation is static.

\begin{figure}[h]
\centering
\includegraphics[width=\textwidth]{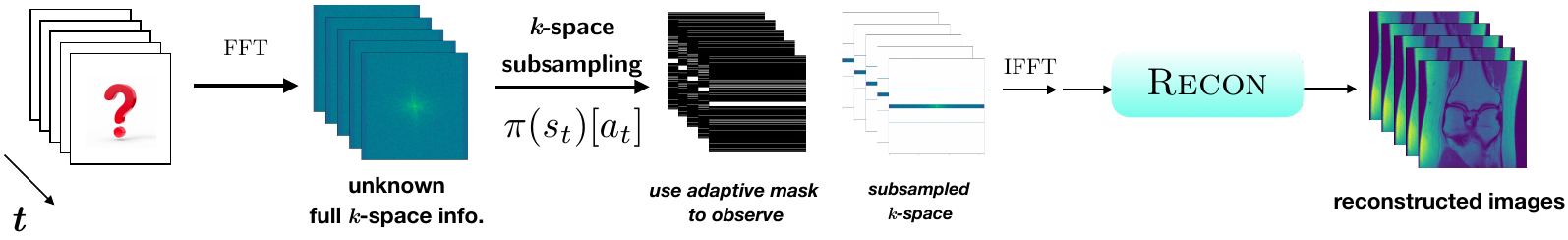}
\caption{Experiment pipeline: sampling and reconstruction}
\end{figure}

We leverage deep Q-learning~\cite{DDQN_2016} and policy gradient method~\cite{RL_SuttonBarto} respectively to show that a sampling policy for collecting limited measurements in the $k$-space adapted
to the underlying image reconstruction method and training dataset can be learned. The major novelty of this work is that the sampling policy is built with respect to a time series of images (such as a serial heartbeat) instead of a set of static images. We also point out that this study is mostly a proof of concept that in the MRI application, RL is effective in learning the sampling strategy that optimizes the performance of the reconstructor (the environment).

\subsection{Related work} Early attempts to learn data-adaptive sampling patterns mostly revolve around static target objects and the parameters of the sampling pattern are
jointly updated along with those of reconstructors. \cite{zhang_cvpr2019} considers one-dimensional Cartesian masks and uses a neural network to evaluate the importance of each unseen row in the frequency domain, so that at inference time, the evaluator network selects the row with highest importance score to acquire. \cite{Huijben_icassp_2020} encodes sampling patterns in a set of distributions, each of which shares the same dimension as the subsampled dimension and is parametrized by a logit. A categorical distribution governs the applicability of each sampling distribution, and all logits and the categorical distribution hyperparameters are co-trained with unrolled ISTA reconstructors~\cite{ISTA_unroll}. This is similar to the LOUPE approach~\cite{LOUPE_2020}, where the sampling mask is directly parameterized as Bernoulli random variables, and \cite{Weiss_icassp_2020}, which uses a careful thresholding operation to handle gradient updates on parameters characterizing one-dimensional Cartesian masks. \cite{Huang_SODUR_2020} introduces an alternating training framework to co-train the reconstructor and a mask predictor to achieve object-level adaptiveness with more explicit control over
the mask predictor training process.

The sequential decision making approach has been adopted to solve the optimal sampling problem in recent studies. \cite{Bakker_policyGradient_nips_2020} explore the policy gradient method to train a policy network that guides a one-dimensional sampling choice with respect to pre-trained reconstructors which are tuned to denoise aliased images blurred by uniformly random sampling masks. \cite{pineda2020activemri} have applied  double Q-learning on constructing adaptive sampling policies for static magnetic resonance images, and the reconstructor in their study is pre-trained in the same manner as in \cite{Bakker_policyGradient_nips_2020}. \cite{Jin_2019_TreeSearch} leverages Monte Carlo tree search to explore mask options, and trains the policy network with tree search selections. Meanwhile, the reconstructor network is alternatively updated along with the
policy network in the training process.

\section{Methodology}
We formalize the sequential selection of the sampling mask for MRI as a partially observable Markov decision process (POMDP). The state at acquisition step $t$ of this POMDP corresponds to the tuple consisting of recent $\alpha$ reconstruction results $s_t = (\widehat{\x}_{t},\cdots, \widehat{\x}_{t-\alpha})$ and the history of actions (i.e., the current sampling mask $\mask_{t-1}$ for the time series of images). An action $a_t$ is to select a particular $k$-space frequency (row or column) to observe. After an action is taken, the sampling mask $\mask_t$ underlying the measurement operation $\samp_t$ has the $a_t$-th diagonal element changed from 0 to 1 compared to $\mask_{t-1}$ for $\samp_{t-1}$.  The reward at each acquisition step is $r_t(s_t, a_t) = \max\big(\mathcal{L}(\recon\circ\samp_t\circ\fft[\x_t],\x_t) - \mathcal{L}(\recon\circ\samp_{t-1}\circ\fft[\x_t],\x_t) , 0\big)$. 

We consider two RL algorithms to build policies in this work: deep double Q-learning and the policy gradient method. Both of these two policies are motivated from the Bellman optimality condition while taking different approximations.
\subsection{Deep double Q-learning (DDQN)} In deep Q-learning, we use a neural network as an evaluating function to determine the value of each action given the current state. A greedy policy ensues by selecting the action with the largest estimated value in the action space. 
The double Q-learning framework keeps two copies of the Q-network during the training process: one $Q_{\param}$ for online action selection while exploring new trajectories, the other, the target network $Q_{\textrm{eval}}$, is the same as the online network except that its parameters are copied every $\tau$ steps from the online network and kept fixed on all other steps. 

We have used the experience replay technique to store observed state transitions and corresponding rewards in memory $\mathcal{M}$. The target network is used for offline evaluation of memorized states.
The training of deep Q learning is set up as a self-supervised problem where the loss function is to minimize the temporal difference error over data sampled from a replay memory buffer. Specifically, the loss function to minimize is formulated as
\begin{align}
&L(\param) = \sum_{\substack{(r_{t+1}, s_{t+1}, s_t, a_t)\\ \in \mathcal{M}}} \big\| \underbrace{R_{t+1} + \gamma Q_{\textrm{eval}}(s_{t+1})[a_{\mathrm{eval}}^\ast] }_{\textrm{target}} -  \underbrace{Q_{\param} (s_t)[a_t]}_{\textrm{online}}  \big\|_1\\
& \textrm{s.t. } a_{\mathrm{eval}}^\ast = \mathrm{arg} \max_a Q_{\textrm{eval}}(s_{t+1})[a]   \nonumber
\end{align}


\subsection{Policy gradient method} Policy gradient methods aim to directly maximize the expected return of a policy parametrized by $\param$ on the POMDP. In this study, we use the REINFORCE algorithm~\cite{RL_SuttonBarto} to train a deep policy network that maximizes the expected sum of rewards given the designated sampling budget. 

In the REINFORCE algorithm, we first collect a whole sequence of state-action pairs up to the time horizon $T$ in a sampling episode, then compute the Q-value $q(s,a)$ based on the partial aggregation of observed state-action data, and finally construct a gradient-based update to policy parameters $\boldsymbol{\theta}$. With the log-ratio trick~\cite{RL_SuttonBarto}, we maximize the following utility functions:
\begin{align}
    J(\param) &= \sum_{t=0}^{T-1} \log \pi_{\param}(s_t)[a_t] \cdot G_t 
    =  \sum_{t=0}^{T-1} \log \pi_{\param}(s_t)[a_t] \cdot \bigg( \sum_{k=t+1}^T \gamma^{k-1} r_k \bigg),
\end{align}
where $a_t = \mathrm{arg}\max_{a} \pi_{\param}(s_t)$
.

\section{Experiments}
\subsection{Dataset} 
We test the RL sampling schemes on the 
\href{https://ocmr.info/}{OCMR data}~\cite{OCMR_arxiv} 
which 
is an open cardiovascular MRI dataset containing multiple cardiac cine series.
We use 62 series for training, and 32 series for testing. Each series in the selected OCMR data is fully sampled, and 
every frame 
is
dimension $384 \times 144$. We aggregate all coils so that the dataset in this study can be considered as single-coiled. The length of each series in the training set varies between 16 and 28, and in the testing set between 16 and 27. There are in total 1519 frames in the training time series and 642 frames in the testing time series.

\subsection{Pre-trained reconstructors}
A modest but viable option of $\recon$ is the Unet~\cite{unet}, which was originally proposed for image segmentation but later was shown to be useful for reconstruction/de-aliasing purposes. To train a Unet in a supervised way as a reconstructor, one needs to set up a training dataset where the aliased images are generated in a certain way. As mentioned in the Related Work section, precedent work on building sampling policies relies on pre-trained reconstructors, where the aliased images in the training dataset are 
subsampled by uniformly random masks in the $k$-space.

In this work, we consider 1D sampling strategies in experiments. We define the \textit{uniformly random} mask as a sampling pattern where the lowest $b$ frequencies in the dimension to be subsampled are always collected while the rest of the sampling budget are uniformly at random allocated across the unobserved high-frequency domain. Here $b$ is a hyperparameter, given the sampling budget. We define the \textit{low-frequency} masks as the sampling pattern where all sampling budget is allocated to the lowest frequencies in the dimension to be subsampled. We define the \textit{energy-distribution-based} masks as the sampling pattern where the lowest $b$ frequencies are always sampled, while the rest of the sampling budget are used according to the energy distribution of each high frequency in the training images.

Before getting into the policy building step, we train three
Unet reconstructors with respect to three separate dataset whose aliased images are generated by each of the aforementioned
masks respectively. Every training image pair in the dataset is in the format of (aliased image $\widetilde{\x}_i^{\boldsymbol{\Xi}}$, corresponding ground truth $\x_i$), where $\widetilde{\x}_i^{\boldsymbol{\Xi}} = \textrm{IFFT} \circ \samp_{\boldsymbol{\Xi}} \circ \textrm{FFT} [\x_i] $. Each pre-trained Unet reconstructor intends to de-aliase the crude IFFT images blurred by a particular sampling scheme $\boldsymbol{\Xi}$: $\recon_{\boldsymbol{\Xi}}(\widetilde{\x}_i^{\boldsymbol{\Xi}})$. We point out that unlike static dataset where energy-distribution-based masks usually outperforms uniformly random masks and low-frequency masks~\cite{Huang_SODUR_2020}, with respect to the OCMR dataset, the pre-trained reconstructors perform the best with the low-frequency mask 
in terms of  both normalized rooted mean squared error (NRMSE) and the structural similarity index measure (SSIM)~\cite{SSIM} as Table~\ref{tab::recon_res} shows.
\begin{table}[H]
\centering
 \begin{tabular}{lll}
 \toprule
 Subsampling schemes $\boldsymbol{\Xi}$ & NRMSE$:=\frac{\|\x_{\textrm{recon}}-\x_\ast\|_2}{\|\x_\ast\|_2}$ & SSIM\\
 \midrule
 Uniformly random masks   & 0.6884 & 0.4380    \\
 Low-frequency masks & \textbf{0.3668} & \textbf{0.7456} \\
 Energy-distribution-based masks & 0.6323 & 0.4816 \\
 \bottomrule
 \\
\end{tabular} 
\caption{Performances of pre-trained reconstructors with respect to aliased images which undergo different sampling schemes. The acceleration ratio is 6-fold with the $b=8$ lowest base frequencies to start and 16 high frequencies to sample. The full size of the subsampled image dimension is 144.}
\label{tab::recon_res}
\end{table}


\subsection{Problem setup}
In this work, we aim to solve the following optimization problem on input time series data $\{\x_i\}_{i=1}^T$:
\begin{align}
\label{eqn::opt_prob_spec}
    &\min_{\mathbf{m}}\sum_{i=1}^T \mathcal{L}( \recon_{\boldsymbol{\Xi}} \circ \samp_{\mathrm{diag}(\mathbf{m})} \circ \fft[\x_i] \, , \, \x_i) \nonumber\\ 
    &\textrm{s.t. } \begin{cases}
    \samp_{\mathrm{diag}(\mathbf{m})} [\y] = \y \cdot \mathrm{diag}(\mathbf{m})  \\
    \sum_{i=1}^n m_i \le b + s \\
    m_i \in \{0,1\}
    \end{cases}
\end{align}
where $\recon_{\boldsymbol{\Xi}}$ is a pretrained Unet with respect to mask pattern $\boldsymbol{\Xi}$ listed in Table \ref{tab::recon_res},  
$s$ is a pre-designated sampling budget, $\y = \fft[\x_i]$ and $\y\in\C^{m\times n}$, $\textrm{diag}(\mathbf{m})$ is an $n\times n$ diagonal matrix with vector $\mathbf{m}$ in its diagonal, and the loss function is $\mathcal{L}(\x,\x_\ast) = \frac{\|\x-\x_\ast\|_2}{\|\x_\ast\|_2}$.

\subsection{Hyperparameter settings}
\paragraph{Acceleration ratio} In accelerated MRI problem, we define the \textit{acceleration ratio} as $\newline\frac{\textrm{the full size of $k$-space dimension to be subsampled}}{\textrm{the actual amount of frequencies to observe}}$.
In this study, we focus on $6$-fold acceleration case, where there are 8 lowest base frequencies to start with in the observation set and we continue to sample additional 16 high frequencies to complete an episode of measurement before 
applying the reconstructor. The total size of the full $k$-space is $384\times 144$, where the second dimension is to be sampled. 

\paragraph{Pre-trained reconstructors} The Unet we use as reconstructor has 6 downsampling and upsampling layers, and has two channels of input, each of which corresponds to the real and imaginary part of the data collected in the Fourier domain
. The loss function we use to train the Unet is $\frac{\|\x_{\textrm{recon}} - \x_{\ast}\|_2}{\|\x_{\ast}\|_2} + 5 \cdot \big(1-\textrm{SSIM}(\x_{\textrm{recon}},\x_\ast)\big)$. We collect each image slice in the time series as the ground truth in the training dataset for sampling policy building, and apply various sampling patterns to generate corresponding aliased images. For reconstructor tuning, with respect to each training dataset generated by different sampling patterns $\boldsymbol{\Xi}$, we train the Unet for 300 epochs, with learning rate as $10^{-5}$, $\ell_2$ regularization weight $10^{-5}$ in the Adam optimizer, and the batchsize 10.

\paragraph{Policy network} We use a convolutional neural network coupled with fully connected feed-forward layers as the policy network. The policy network sees the reconstruction images in the previous 8 time steps, and output the probability distribution on the remaining unobserved high frequencies to sample. 

\paragraph{DDQN algorithm setup} We set the size of the memory buffer as 100. We use the Adam optimizer with $10^{-5}$ as the constant learning rate. We run DDQN training for 600 epochs with the discount factor set as 0.8 and batch size 3. 

\paragraph{REINFORCE algorithm setup} We add an \textit{entropy regularization} term~\cite{entropy_reg_RL_2016} to the loss function in order to encourage the learning process to explore larger trajectory space, and we set the weight of this penalty term to be 0.01. We use the Adam optimizer with $10^{-5}$ as the constant learning rate. We run REINFORCE training for 600 epochs with discount factor set as 0.1 and batch size 3.

\subsection{Validation for the proof of concept}
\subsubsection{Convergence results} 
\begin{figure}[h]
\centering
\begin{subfigure}{.5\textwidth}
  \centering
  \includegraphics[width=.9\linewidth]{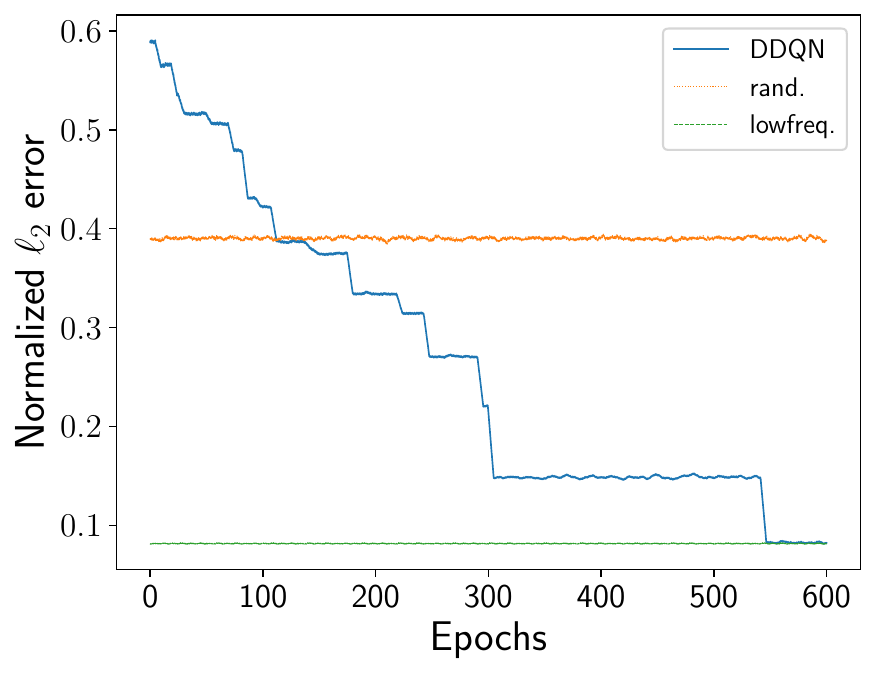}
  \caption{DDQN RMSE convergence curve}
  \label{fig::DDQN_RMSE}
\end{subfigure}%
\begin{subfigure}{.5\textwidth}
  \centering
  \includegraphics[width=.9\linewidth]{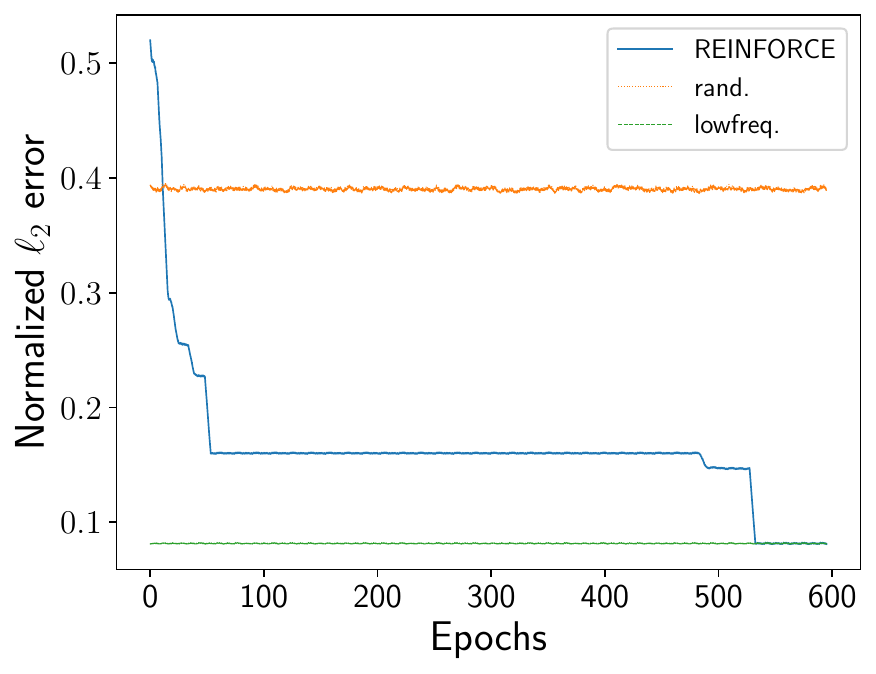}
  \caption{REINFORCE RMSE convergence curve}
    \label{fig::REINFORCE_RMSE}
\end{subfigure}
\caption{Convergence curve of image reconstruction error (normalized $\ell_2$ error) during training.}
\label{fig:test}
\end{figure}

The following figures show the convergence of the reinforcement learning training paradigm. Figures \ref{fig::DDQN_RMSE} and \ref{fig::REINFORCE_RMSE} show that the masks predicted by the policy network trained by DDQN and REINFORCE, respectively, demonstrate improved performance in terms of the image reconstruction error compared to the fixed pre-trained reconstructor. Both DDQN and REINFORCE can recover the sampling scheme under which the reconstructor is trained. In this case, the underlying sampling scheme is low-frequency sampling policy.

\begin{figure}[h]
\centering
\includegraphics[width=\textwidth]{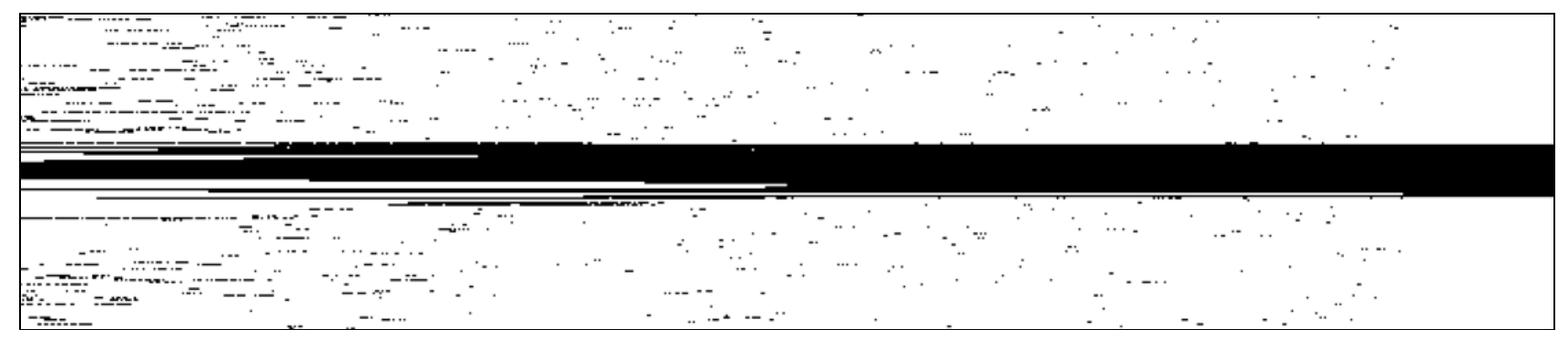}
\caption{Mask evolution history as DDQN training proceeds. The horizontal axis shows the direction of time, 
and the vertical axis is the frequency grid of the subsampled dimension in $k$-space. We put the low frequency in the middle of the vertical axis and the high frequencies on the upper and lower end of the vertical axis. Each vertical cross section is a sampling mask at a given iteration step returned by the policy network trained through DDQN algorithm. }
\label{fig::DDQN_mask}
\end{figure}

\begin{figure}[h]
\centering
\includegraphics[width=\textwidth]{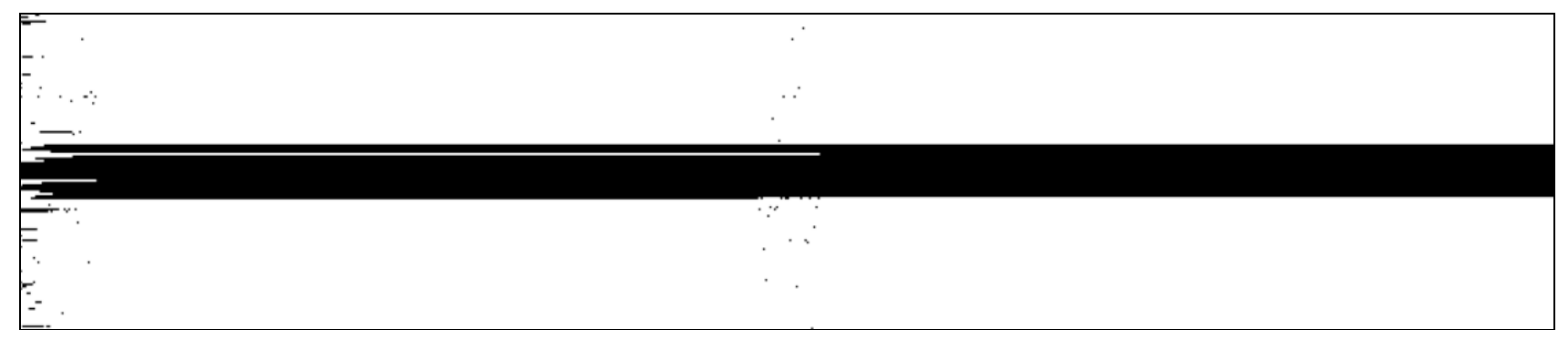}
\caption{Mask evolution history as REINFORCE training proceeds. The horizontal axis shows the direction of time elapses, and the vertical axis is the frequency grid of the subsampled dimension in $k$-space. We put the low frequency in the middle of the vertical axis and the high frequencies on the upper and lower end of the vertical axis. Each vertical cross section is a sampling mask at a given iteration step returned by the policy network trained through REINFORCE algorithm.}
\label{fig::REINFORCE_mask}
\end{figure}

Figures \ref{fig::DDQN_mask} and \ref{fig::REINFORCE_mask} show the evolution of predicted masks returned by the policy network. Both policies converge to the low-frequency sampling policy, while the DDQN paradigm takes more iterations to discover ultimate target.

\begin{table}[H]
\centering
 \begin{tabular}{lll}
 \toprule
 Subsampling schemes/Policies & NRMSE$:=\frac{\|\x_{\textrm{recon}}-\x_\ast\|_2}{\|\x_\ast\|_2}$ & SSIM\\
 \midrule
 DDQN trained policy & 0.3948 & \textbf{0.7522} \\
 REINFORCE trained policy & \textbf{0.3943} & 0.7521\\
 Uniformly random masks   & 0.5638 & 0.5793    \\
 Low-frequency masks & 0.3948 & \textbf{0.7522} \\
 Energy-based masks & 0.6160 & 0.4825 \\
 \bottomrule
 \\
\end{tabular} 
\caption{Reconstruction accuracy on the testing dataset by different policies. Note that the set of testing images are different from that in Table \ref{tab::recon_res} in the sense that only frames in time series after the initial base observations are taken into account. In other words, for each time series in a test file, we consider reconstructions on frames $\x_{\textrm{base}
}+1,\cdots, \x_{\textrm{end}}$ where ``base" is the count of base frequencies we use as basic observations. } 
\end{table}

\subsubsection{Reconstruction examples}
\begin{figure}
\centering
\includegraphics[width=\textwidth]{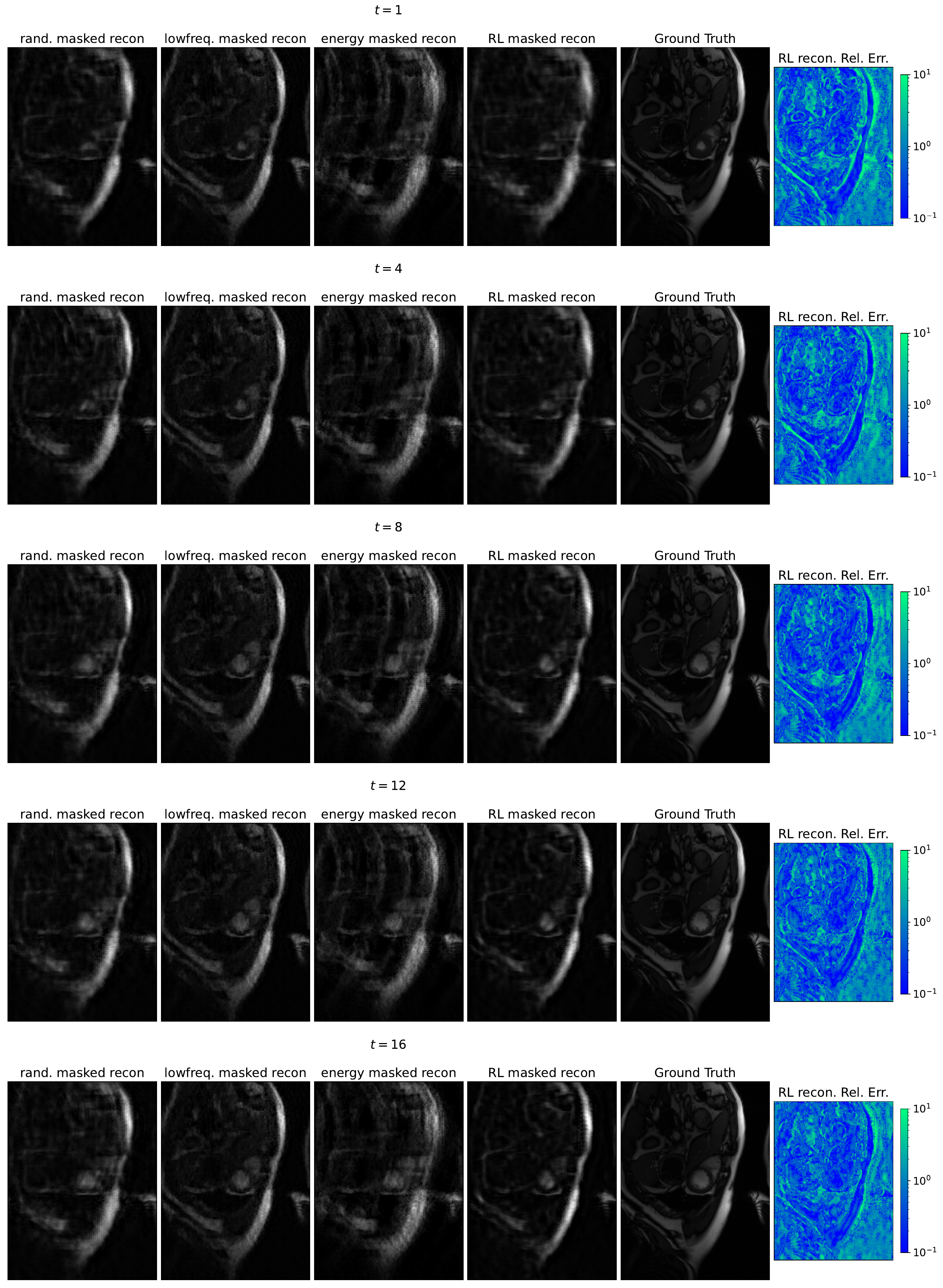}
\caption{An example for a cardiac cycle that contains reconstruction (left column), ground truth (central column) and error maps (right column).}
\label{fig::REINFORCE_eg}
\end{figure}

In Figure \ref{fig::REINFORCE_eg}, we show a set of reconstruction examples in the testing set as sampling proceeds. The time series characterizes a cardiac cycle where the diastole is followed by the systole. We show the ground truth images in the cardiac cycle, their corresponding reconstruction images, and error maps.  When generating reconstructions on the testing time sequences, we first assemble the sampling masks according to policies under comparison and the designated sampling budget before applying corresponding reconstructors. After masks are generated regarding a testing time series, they remain fixed for the rest of frames in that time series. The sampling masks used for measurement in $k$-space starts from the base $b=8$ low frequencies, and garner additional $s=16$ high frequencies according to sampling policies. The mask used for each frame in Figure \ref{fig::REINFORCE_eg} consists of both base low frequencies and policy-informed high frequencies.

\section{Conclusion}
In this work, we present a synthetic study where with prior knowledge on the performance of candidate reconstructors,
reinforcement learning techniques can effectively uncover the underlying mask patterns with which the reconstructor is trained to denoise undersampled dynamic cardiac cine sequences. This study also shows that such dynamic MRI datasets like OCMR data lead to new challenges as the conventional reconstructors and energy-based sampling schemes do not demonstrate the same performance as in the static setting.
\paragraph{Future work}
Several aspects of using RL algorithms can be further investigated:
\begin{enumerate}
    \item How can we accelerate the training process or improve the reconstruction quality through more sophisticated RL optimization practice~\cite{TRPO} or leverage prior information of data distribution in the training process?
    \item Do reconstructors such as MoDL~\cite{MoDL} or unrolling based reconstructors~\cite{DL_for_InverseImaging,modelBased_DL} with higher complexity improve the reconstruction accuracy for the dynamic dataset?
    \item How can we develop a self-supervised training process to either simultaneously or alternatively train the policy network and the reconstructor network such that the final reconstruction performance can outperform the starting baseline?
\end{enumerate}

\paragraph{Acknowledgement}
    Zhishen Huang thanks Prof.~Saiprasad Ravishankar for helpful discussion when Zhishen was with Michigan State University. 

\bibliographystyle{plain}
\bibliography{references}
\end{document}

%% file: header.tex
\usepackage{amsfonts}
\usepackage{amsthm,amsmath,amsfonts,amssymb}
\usepackage{graphicx}
\usepackage[font=small,labelfont=bf]{caption}
\usepackage{epstopdf}
\usepackage{float}
\usepackage{pgfplots}
\usepackage{listings}
\usepackage{longtable}
\usepackage{mathrsfs}
\usepackage{bm}
\usepackage{extarrows}
\usepackage{comment}
\usepackage{booktabs}

\usepackage{cleveref}
\usepackage{bbm}
\usepackage{tikz}

\theoremstyle{definition}

\theoremstyle{remark}

\definecolor{energy}{RGB}{114,0,172}
\definecolor{freq}{RGB}{45,177,93}
\definecolor{spin}{RGB}{251,0,29}
\definecolor{signal}{RGB}{203,23,206}
\definecolor{circle}{RGB}{217,86,16}
\definecolor{average}{RGB}{203,23,206}

\colorlet{shadecolor}{gray!20}
\pgfplotsset{compat=1.9}

\usepgflibrary{fpu}

\newcommand{\R}{\mathbb{R}}
\newcommand{\C}{\mathbb{C}}

\newcommand{\x}{\mathbf{x}}
\newcommand{\y}{\mathbf{y}}

\newcommand{\mask}{\mathbf{M}}
\newcommand{\samp}{\mathcal{S}}
\newcommand{\recon}{\mathcal{G}}

\newcommand{\param}{\boldsymbol{\theta}}

\newcommand{\fft}{\mathcal{F}}


%% file: main_arxiv.bbl
\begin{thebibliography}{10}

\bibitem{MoDL}
Hemant~K. Aggarwal, Merry~P. Mani, and Mathews Jacob.
\newblock {MoDL}: Model-based deep learning architecture for inverse problems.
\newblock {\em IEEE Transactions on Medical Imaging}, 38(2):394--405, 2019.

\bibitem{LOUPE_2020}
Cagla~D. Bahadir, Alan~Q. Wang, Adrian~V. Dalca, and Mert~R. Sabuncu.
\newblock Deep-learning-based optimization of the under-sampling pattern in
  {MRI}.
\newblock {\em IEEE Transactions on Computational Imaging}, 6:1139--1152, 2020.

\bibitem{Bakker_policyGradient_nips_2020}
Tim Bakker, Herke van Hoof, and Max Welling.
\newblock Experimental design for {MRI} by greedy policy search.
\newblock In H.~Larochelle, M.~Ranzato, R.~Hadsell, M.F. Balcan, and H.~Lin,
  editors, {\em Advances in Neural Information Processing Systems}, volume~33,
  pages 18954--18966. Curran Associates, Inc., 2020.

\bibitem{Becker2011PracticalCS}
Stephen Becker.
\newblock {\em Practical Compressed Sensing: Modern data acquisition and signal
  processing}.
\newblock PhD thesis, California Institute of Technology, 2011.

\bibitem{candes_cs}
Emmanuel~J. Candes and Michael~B. Wakin.
\newblock An introduction to compressive sampling.
\newblock {\em IEEE Signal Processing Magazine}, 25(2):21--30, 2008.

\bibitem{OCMR_arxiv}
Chong Chen, Yingmin Liu, Philip Schniter, Matthew Tong, Karolina Zareba,
  Orlando Simonetti, Lee Potter, and Rizwan Ahmad.
\newblock {OCMR} (v1.0)--open-access multi-coil k-space dataset for
  cardiovascular magnetic resonance imaging, 2020.

\bibitem{ISTA_unroll}
Karol Gregor and Yann LeCun.
\newblock Learning fast approximations of sparse coding.
\newblock In {\em ICML 2010 - Proceedings, 27th International Conference on
  Machine Learning}, ICML 2010 - Proceedings, 27th International Conference on
  Machine Learning, pages 399--406, 2010.
\newblock 27th International Conference on Machine Learning, ICML 2010 ;
  Conference date: 21-06-2010 Through 25-06-2010.

\bibitem{Huang_SODUR_2020}
Zhishen Huang and Saiprasad Ravishankar.
\newblock Single-pass object-adaptive data undersampling and reconstruction for
  {MRI}.
\newblock {\em IEEE Transactions on Computational Imaging}, 8:333--345, 2022.

\bibitem{huang2021modelbased}
Zhishen Huang, Siqi Ye, Michael~T. McCann, and Saiprasad Ravishankar.
\newblock Model-based reconstruction with learning: From unsupervised to
  supervised and beyond, 2021.

\bibitem{Huijben_icassp_2020}
Iris~A.M. Huijben, Bastiaan~S. Veeling, and Ruud~J.G. van Sloun.
\newblock Learning sampling and model-based signal recovery for compressed
  sensing {MRI}.
\newblock In {\em ICASSP 2020 - 2020 IEEE International Conference on
  Acoustics, Speech and Signal Processing (ICASSP)}, pages 8906--8910, 2020.

\bibitem{Jin_2019_TreeSearch}
Kyong~Hwan Jin, Michael Unser, and Kwang~Moo Yi.
\newblock Self-supervised deep active accelerated {MRI}.
\newblock {\em CoRR}, abs/1901.04547, 2019.

\bibitem{entropy_reg_RL_2016}
Volodymyr Mnih, Adria~Puigdomenech Badia, Mehdi Mirza, Alex Graves, Timothy
  Lillicrap, Tim Harley, David Silver, and Koray Kavukcuoglu.
\newblock Asynchronous methods for deep reinforcement learning.
\newblock In Maria~Florina Balcan and Kilian~Q. Weinberger, editors, {\em
  Proceedings of The 33rd International Conference on Machine Learning},
  volume~48 of {\em Proceedings of Machine Learning Research}, pages
  1928--1937, New York, New York, USA, 20--22 Jun 2016. PMLR.

\bibitem{unrolling_eldar}
Vishal Monga, Yuelong Li, and Yonina~C. Eldar.
\newblock Algorithm unrolling: Interpretable, efficient deep learning for
  signal and image processing.
\newblock {\em IEEE Signal Processing Magazine}, 38(2):18--44, 2021.

\bibitem{DL_for_InverseImaging}
Gregory Ongie, Ajil Jalal, Christopher~A. Metzler, Richard~G. Baraniuk,
  Alexandros~G. Dimakis, and Rebecca Willett.
\newblock Deep learning techniques for inverse problems in imaging.
\newblock {\em IEEE Journal on Selected Areas in Information Theory},
  1(1):39--56, 2020.

\bibitem{pineda2020activemri}
Luis Pineda, Sumana Basu, Adriana Romero, Roberto Calandra, and Michal
  Drozdzal.
\newblock Active {MR} k-space sampling with reinforcement learning.
\newblock In Anne~L. Martel, Purang Abolmaesumi, Danail Stoyanov, Diana Mateus,
  Maria~A. Zuluaga, S.~Kevin Zhou, Daniel Racoceanu, and Leo Joskowicz,
  editors, {\em Medical Image Computing and Computer Assisted Intervention --
  MICCAI 2020}, pages 23--33, Cham, 2020. Springer International Publishing.

\bibitem{transform_learning_Sai}
Saiprasad Ravishankar, Bihan Wen, and Yoram Bresler.
\newblock Online sparsifying transform learning—part {I}: Algorithms.
\newblock {\em IEEE Journal of Selected Topics in Signal Processing},
  9(4):625--636, 2015.

\bibitem{Sai_recon_review}
Saiprasad Ravishankar, Jong~Chul Ye, and Jeffrey~A. Fessler.
\newblock Image reconstruction: From sparsity to data-adaptive methods and
  machine learning.
\newblock {\em Proceedings of the IEEE}, 108(1):86--109, 2020.

\bibitem{unet}
Olaf Ronneberger, Philipp Fischer, and Thomas Brox.
\newblock U-net: Convolutional networks for biomedical image segmentation.
\newblock In Nassir Navab, Joachim Hornegger, William~M. Wells, and
  Alejandro~F. Frangi, editors, {\em Medical Image Computing and
  Computer-Assisted Intervention -- MICCAI 2015}, pages 234--241, Cham, 2015.
  Springer International Publishing.

\bibitem{TRPO}
John Schulman, Sergey Levine, Pieter Abbeel, Michael Jordan, and Philipp
  Moritz.
\newblock Trust region policy optimization.
\newblock In Francis Bach and David Blei, editors, {\em Proceedings of the 32nd
  International Conference on Machine Learning}, volume~37 of {\em Proceedings
  of Machine Learning Research}, pages 1889--1897, Lille, France, 07--09 Jul
  2015. PMLR.

\bibitem{modelBased_DL}
Nir Shlezinger, Jay Whang, Yonina~C. Eldar, and Alexandros~G. Dimakis.
\newblock Model-based deep learning.
\newblock {\em Proceedings of the IEEE}, pages 1--35, 2023.

\bibitem{RL_SuttonBarto}
Richard~S. Sutton and Andrew~G. Barto.
\newblock {\em Reinforcement Learning: An Introduction}.
\newblock A Bradford Book, Cambridge, MA, USA, 2018.

\bibitem{DDQN_2016}
Hado van Hasselt, Arthur Guez, and David Silver.
\newblock Deep reinforcement learning with double q-learning.
\newblock {\em Proceedings of the AAAI Conference on Artificial Intelligence},
  30(1), Mar. 2016.

\bibitem{SSIM}
Zhou Wang, A.C. Bovik, H.R. Sheikh, and E.P. Simoncelli.
\newblock Image quality assessment: from error visibility to structural
  similarity.
\newblock {\em IEEE Transactions on Image Processing}, 13(4):600--612, 2004.

\bibitem{Weiss_icassp_2020}
Tomer Weiss, Sanketh Vedula, Ortal Senouf, Oleg Michailovich, Michael
  Zibulevsky, and Alex Bronstein.
\newblock Joint learning of cartesian under sampling andre construction for
  accelerated {MRI}.
\newblock In {\em ICASSP 2020 - 2020 IEEE International Conference on
  Acoustics, Speech and Signal Processing (ICASSP)}, pages 8653--8657, 2020.

\bibitem{zhang_cvpr2019}
Zizhao Zhang, Adriana Romero, Matthew~J. Muckley, Pascal Vincent, Lin Yang, and
  Michal Drozdzal.
\newblock Reducing uncertainty in undersampled {MRI} reconstruction with active
  acquisition.
\newblock In {\em 2019 IEEE/CVF Conference on Computer Vision and Pattern
  Recognition (CVPR)}, pages 2049--2053, 2019.

\end{thebibliography}
